\documentclass[runningheads]{llncs}
\usepackage[T1]{fontenc}
\usepackage{graphicx}
\usepackage{amsfonts}
\usepackage{amsmath}
\usepackage{amssymb}
\usepackage{bm}
\usepackage{tabularx}
\usepackage{multirow}
\usepackage{subfig}
\usepackage{tikz}
\usepackage{xspace}
\usepackage[acronym, shortcuts]{glossaries}
\usepackage{colortbl}
\usepackage{tcolorbox}
\usepackage{url}

\newcommand{\best}[1]{\pdfliteral direct {2 Tr 0.3 w}#1\pdfliteral direct {0 Tr 0 w}}

\newcolumntype{L}{>{\raggedright\arraybackslash}X}
\newcolumntype{C}{>{\centering\arraybackslash}X}
\newcolumntype{R}{>{\raggedleft\arraybackslash}X}

\renewcommand{\vec}[1]{\mbox{\textbf{#1}} }

\newacronym{ctd}{ConcTD}{Concrete Temporal Dropout}
\newacronym{td}{TD}{Temporal Dropout}
\newacronym{cd}{ConcD}{Concrete Dropout}
\newacronym{eo}{EO}{Earth Observation}
\newacronym{rs}{RS}{Remote Sensing}
\newacronym{dl}{DL}{Deep Learning}
\newacronym{mlp}{MLP}{Multi-layer Perceptron}
\newacronym{mcctd}{MC-ConcTD}{Monte Carlo Concrete Temporal Dropout}
\newacronym{mctd}{MC-TD}{Monte Carlo Temporal Dropout}
\newacronym{mc}{MC}{Monte Carlo}
\newacronym{bnn}{BNN}{Bayesian Neural Network}
\newacronym{nll}{NLL}{Negative Log-Likelihood}
\newacronym{lstm}{LSTM}{Long-Short Term Memory}
\newacronym{rnn}{RNN}{Recurrent Neural Network}
\newacronym{cnn}{CNN}{Convolutional Neural Network}
\newacronym{vi}{VI}{Variatioinal Inference}

\newacronym{uq}{UQ}{Uncertainty Quantification}

\newacronym{yield}{SwissYield}{Swiss Crop Yield}
\newacronym{lfmc}{LFMC}{Live Fuel Moisture Content}
\newacronym{pm25}{PM2.5}{Particles Matter 2.5}

\begin{document} 
\title{An Analysis of Temporal Dropout in Earth Observation Time Series for Regression Tasks}
\titlerunning{Temporal Dropout for EO Time Series Data}
%
\author{Miro Miranda\inst{1,2} \orcidID{0009-0002-8195-9776} \and
Francisco Mena\inst{1,2}\orcidID{0000-0002-5004-6571} \and
Andreas Dengel\inst{1,2}\orcidID{0000-0002-6100-8255}}
\authorrunning{M. Miranda et al.}
%
\institute{University of Kaiserslautern-Landau, Kaiserslautern, Germany \and
German Research Center for Artificial Intelligence, Kaiserslautern, Germany \\
\email{\{miro.miranda\_lorenz,francisco.mena,andreas.dengel\}@dfki.de}\\
}

\maketitle              

\begin{abstract} 
Missing instances in time series data impose a significant challenge to deep learning models, particularly in regression tasks. 
In the Earth Observation field, satellite failure or cloud occlusion frequently results in missing time-steps, introducing uncertainties in the predicted output and causing a decline in predictive performance. 
While many studies address missing time-steps through data augmentation to improve model robustness, the uncertainty arising at the input level is commonly overlooked.
To address this gap, we introduce Monte Carlo Temporal Dropout (MC-TD), a method that explicitly accounts for input-level uncertainty by randomly dropping time-steps during inference using a predefined dropout ratio, thereby simulating the effect of missing data. To bypass the need for costly searches for the optimal dropout ratio, we extend this approach with Monte Carlo Concrete Temporal Dropout (MC-ConcTD), a method that learns the optimal dropout distribution directly.
Both MC-TD and MC-ConcTD are applied during inference, leveraging Monte Carlo sampling for uncertainty quantification.
Experiments on three EO time-series datasets demonstrate that MC-ConcTD improves predictive performance and uncertainty calibration compared to existing approaches. Additionally, we highlight the advantages of adaptive dropout tuning over manual selection, making uncertainty quantification more robust and accessible for EO applications.

\end{abstract}

\section{Introduction} \label{sec:introduction}
Recently, \gls{dl} models have been widely used in the \gls{eo} field to find optimal data-driven solutions in different applications \cite{camps2021deep}.
\gls{dl} effectively processes complex and heterogeneous sensor data, allowing for accurate analysis of environmental patterns \cite{mena2024common}.
For instance, predicting continuous values such as crop yield \cite{perich2023pixel}, water content in plants \cite{rao2020sar}, and surface forecast \cite{requena2021earthnet2021} involves processing complex data. 
In the \gls{eo} field, processing time series sensor data is essential for understanding the changes and dynamics of our planet \cite{miller2024deep}. 
However, sensors may experience anomalies and occlusions, leading to missing data over certain time-steps \cite{shen2015missing,garnot2022multi,ebel2023uncrtaints}.
For instance, clouds obstruct the sunlight in optical images, as on average, 67\% of Earth's surface is covered by clouds \cite{king2013spatial}, often resulting in uncertain predicted outputs. 
Addressing missing data in time series is a prevalent challenge in the \gls{dl} field, with various modeling solutions and pre-processing techniques developed to mitigate its impact and ensure accurate predictions.  
While various techniques exist to mitigate missing data \cite{garnot2022multi,mena2024increasing}, few explicitly quantify the uncertainty it introduces, particularly the impact of input-level uncertainty. Addressing this gap is crucial for improving model reliability in EO regression applications.

In this work, we analyze two models that simulate missing data across time series during training and inference for a dual purpose. 
During training, it acts as an augmentation technique to increase model generalization to missing data in temporal \gls{eo} data.
During inference, by generating multiple \gls{mc} samples with different missing patterns, it acts as an \gls{uq} mechanism, thereby increasing the trustworthiness of a prediction. 
The first model is built on the dropout variational distribution applied during inference \cite{gal2016dropout} and over time, referred to as \gls{mctd}.
However, the optimal dropout value is a difficult hyperparameter to tune, requiring a resource-intensive process that depends on each dataset and missing data type \cite{mena2024increasing}. 
Instead of manually searching, we propose using \gls{cd} \cite{gal2017concrete} to learn the optimal dropout ratio using standard gradient descent. We refer to this method as \gls{mcctd}, where \gls{cd} is applied across time and during inference.
The \gls{cd} \cite{gal2017concrete} approximates a Bernoulli distribution using a continuous relaxation, enabling the dropout probability to be optimized through gradient-based learning \cite{maddison2017concrete}.
Unlike traditional dropout-based approaches, this model automatically learns an optimal dropout distribution, eliminating costly hyperparameter tuning. 

We validate \gls{mctd} and \gls{mcctd} on regression tasks spanning three \gls{eo} datasets with temporal sensor data.
Experimental results demonstrate that \gls{mcctd} improves both predictive accuracy and uncertainty calibration, while also eliminating the need for manual dropout tuning. The \gls{ctd} not only enhances predictive performance but also improves the accessibility and practicality of \gls{uq} in \gls{eo} regression applications.
The code is publicly available at \url{https://github.com/mmiranda-l/Temporal-Dropout/}.
\section{Related Work} \label{sec:related}

\paragraph{Missing data in time series.}
Irregular sampled time series are common in signal processing \cite{orfanidis1995introduction}. 
Temporal observations can easily suffer from problems in their collection, causing irregular observations with missing data.
Numerous research in \gls{dl} has focused on learning to impute time series, such as BRITS \cite{cao2018brits}, mTAN \cite{shuklamulti}, and SAITS \cite{du2023saits}, while others focused on adapting models to ignore the missing data, such as D-GRU \cite{che2018recurrent} and MissFormer \cite{becker2021missformer}.
In the \gls{eo} field, missing spatial and temporal data are a common phenomenon due to real-world data collection constraints \cite{shen2015missing}. This includes, sensor noise, sensor failure, and cloud occlusion, affecting observations and introducing uncertainty. 
This problem negatively affects the performance of predictive models, where more missing data translates to worse predictions \cite{inglada2016improved,ferrari2023fusing}.
However, some strategies in the \gls{eo} field mitigate the negative effect of missing data, such as including features from different sensors or using dropout techniques \cite{ofori2021crop,garnot2022multi}. 
Recently, the \gls{td} technique, which involves randomly dropping time steps, has been used to enhance the model performance \cite{garnot2022multi,tseng2023lightweight,helber2024operational}.
Furthermore, some studies leverage missing data as an augmentation technique to enhance generalization. For instance, Weitland et al. \cite{weilandt2023early} randomly mask the end of a time-series for generalization to an early crop classification.
On the other hand, some studies have focused on reconstruction tasks that recover the missing time-steps. 
For instance, Chen et al. \cite{chen2004simple} use a polynomial fit based on the Savitzky–Golay filter, while others use \gls{dl} models based on \gls{mlp} \cite{das2017deep}, or convolutions \cite{scarpa2018cnn}.

\paragraph{Regression with time series data.}
Regression tasks have been lesser explored than classification tasks with time-series data \cite{mohammadi2024deep}.
One reason might be that \gls{dl} models are not as effective for regression as for classification tasks, as shown by Tan et al. \cite{tan2021time}. The intrinsic challenge of predicting a continuous value instead of a categorical value is often overlooked. Nevertheless, in the \gls{eo} field, there are numerous applications involving pixel-level regression with time series data.
For instance, Nguyen et al. \cite{nguyen2019spatial} use multi-spectral data and weather time series for pixel-wise crop yield prediction. They use a \gls{mlp} model for processing the time series data as individual features. However, it has been shown recently that \gls{rnn} models obtain better results. This is underlined by \cite{pathak2023predicting,miranda2024multi}.
Furthermore, Maimaitijiang et al. \cite{maimaitijiang2020soybean} consider the use of drone-based optical time series data for the same task, obtaining images unaffected by cloud occlusion.
Another pixel-wise task studied in the literature is cloud removal. 
This task has been explored with multi-spectral optical time series data \cite{sarukkai2020cloud}, as well as including radar time series \cite{ebel2023uncrtaints}.
Similarly, in the Earth surface forecast, multi-spectral optical and weather time series have been used with spatio-temporal models to obtain accurate predictions \cite{requena2021earthnet2021}, such as using ConvLSTM models \cite{diaconu2022understanding}.

\paragraph{Uncertainty estimation.}
Uncertainty estimation holds particular promise in safety-critical applications, including the EO field \cite{gawlikowski2023survey}. Reliable uncertainty estimates improve decision-making by identifying predictions with high confidence and those that should be treated with caution \cite{kendall2017uncertainties,hullermeier2021aleatoric}. Uncertainty in predictive models is typically classified into two categories: epistemic and aleatoric \cite{kendall2017uncertainties,hullermeier2021aleatoric}.
While the former captures the uncertainty coming from the model itself, the latter represents the uncertainty coming from the data exclusively, such as sensor noise. In regression, the aleatoric uncertainty is often directly learned from the data using proper scoring rules such as the Gaussian \gls{nll} loss \cite{nix1994estimating}. In contrast, epistemic uncertainty is typically estimated by modeling a distribution over the weight space, often based on Bayes' Theorem. This includes approximate Bayesian inference \cite{blundell2015weight}, or dropout techniques \cite{gal2016dropout,gal2017concrete,mobiny2021dropconnect}.
In the \gls{eo} field, numerous studies have assessed the effectiveness of \gls{uq} methods, particularly in applications such as crop yield prediction with time series data, as evidenced in \cite{ma2021corn}.
However, most of these studies overlook the uncertainty arising at the input level, which can be crucial in improving model reliability and robustness. 
In our work, 
we explore the simulation of missing data in \gls{eo} time series for regression tasks. 
To the best of our knowledge, there is no method in the \gls{eo} literature that accounts for the uncertainty arising from the input time series using missing time steps.

\section{Temporal Dropout for Uncertainty Estimation} \label{sec:methods}

\subsection{Notation \& Preliminaries}
Let $\vec{x} \in \mathcal{X}$ and $y \in \mathcal{Y}$ denote samples from the input and target spaces, respectively. In the case of multivariate time series data with $T$ time-steps, $\vec{x}$ is given as $\vec{x} = (x_0, ..., x_T)$. We further denote data pairs as $(\vec{x}, y)$, and define the goal to predict the target $y$ from $\vec{x}$. 
Thus, we aim to find a model  $f^\theta(\cdot): \mathcal{X} \longmapsto \mathcal{Y}$ such that $f^{\theta}(\vec{x}) = \hat{y}$, by optimizing over the model's parameters $\theta$. 

\paragraph{Uncertainty estimation:}
We define the problem of learning the uncertainty in the predicted output ($\hat{y} = \{\mu, \sigma^2\}$) propagated through the uncertainty in the input and the model itself. 
To estimate the epistemic uncertainty (EU), we model the expected output and variance over multiple stochastic forward passes, by placing a distribution over the network. Given independent input samples $x^{(l)}$ drawn from the input distribution, EU is approximated as follows: 
\begin{align*}
\mathbb{E}[f^\theta(x)] = \mu_{EU} & \approx L^{-1} \sum_l f^\theta(x^{(l)}) \\    
Var[f^\theta(x)] = \sigma^2_{EU} & \approx  (L-1)^{-1} \sum_l (f^\theta(x^{(l)}) - \mathbb{E}[f^\theta(x)])^2.
\end{align*}
To estimate the aleatoric uncertainty (AU), which arises from the data, we assume a Normal distribution parameterized by two output heads: $\mu (x), \sigma^2 (x) = f^\theta(x)$. Both terms can be optimized by minimizing the \gls{nll} \cite{nix1994estimating}: 
\begin{equation}
\mathcal{L}_{NLL}(y, x) = \log\sigma^2(x) + \frac{(\mu(x) - y)^2}{ \sigma^2(x)}.    
\end{equation} 
This approach captures AU in $\sigma^2(x)$, allowing the model to capture data related uncertainty. 
In Valdenegro-Toro et al. \cite{valdenegro2024unified} a relationship between input and output uncertainty is given, showing that input uncertainty is propagated through the model, finally resulting in model, or epistemic uncertainty (EU) by leveraging stochastic \gls{mc} sampling. We follow this argumentation. Finally, the predictive uncertainty (PU) is the combined effect of all uncertainties.

\subsection{Temporal Dropout}
We leverage \gls{td} to estimate uncertainty. The dropout technique \cite{srivastava2014dropout} is commonly used in hidden layers of \gls{dl} models during training. 
Nevertheless, applying this technique at the input-level across time has shown good regularization performance in the \gls{eo} field \cite{metzger2021crop,garnot2022multi,tseng2023lightweight,mena2024increasing,helber2024operational}. 
This prevents the model from focusing on individual time-steps and simultaneously handling missing time-steps, which randomly occur in \gls{eo} data with sensor failure and cloud coverage.
Thus, the multivariate time series data $\vec{x}$ is masked out as
\begin{equation}
    \hat{\vec{x}} = \vec{x} \odot (1-\vec{p}),
\end{equation}
where $\vec{p} \in [0,1]^T$ is the binary dropout mask, which is drawn from a Bernoulli distribution, i.e. $p_i \sim \text{Bern}(\alpha), \forall i = 1,...,T$, with $\alpha$ as the dropout ratio. 
Thus, the input $\vec{x}$ becomes a random variable $\hat{\vec{x}}$, providing additional network regularization. In contrast, we estimate the uncertainty arising from missing time steps, by enabling  \gls{td} at the inference level. Gal et al. \cite{gal2016dropout} demonstrated that casting dropout approximates Bayesian inference in any neural network architecture. Similarly, we extend this formulation along the temporal dimension, named \textbf{\acrfull{mctd}}. 
This corresponds to sampling $L$ different dropout masks for prediction, $p_i^{(l)} \sim \text{Bern}(\alpha)$, $\forall l = 1, \ldots, L$, and applying them to the input data as $\hat{\vec{x}}^{(l)} = \vec{x} \odot (1-\vec{p}^{(l)})$.

\subsection{Concrete Temporal Dropout}
\begin{figure*}[t]
\centering
\includegraphics[width=.85\textwidth, page=1]{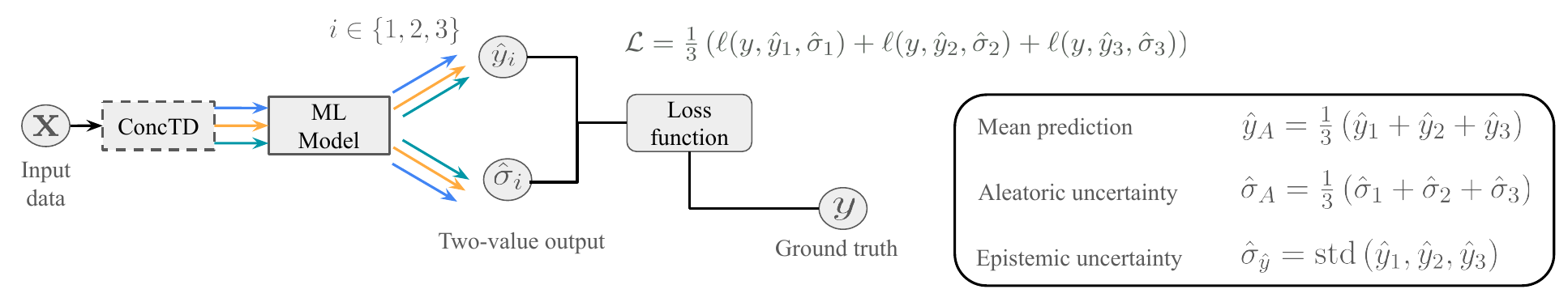}
\caption{Illustration of a DL model prediction with the \gls{ctd} technique and three \gls{mc} samples indexed by $i \in \{1,2,3\}$.} \label{fig:training}
\end{figure*}

\begin{figure*}[!t]
\centering
\includegraphics[width=0.85\textwidth, page=1]{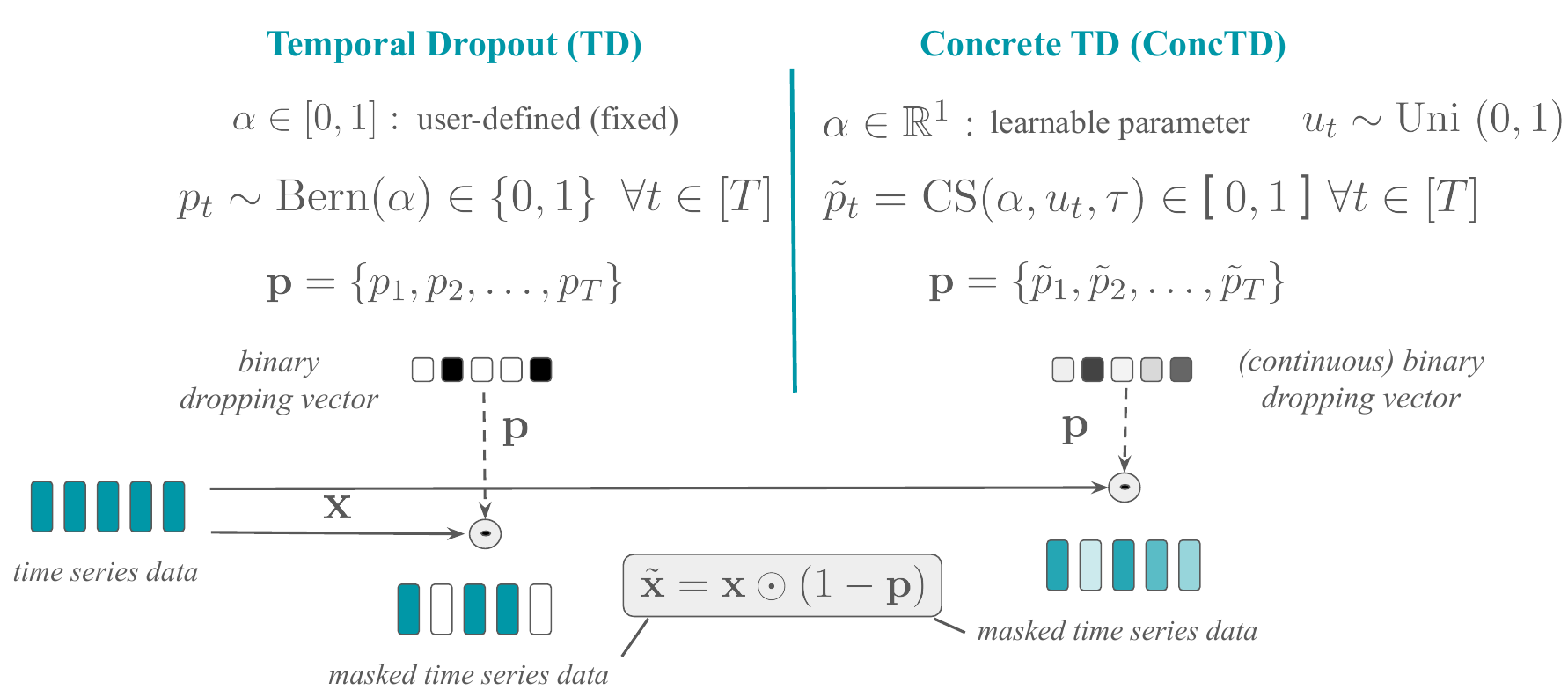}
\caption{Illustration of \gls{td} and \gls{ctd} techniques over time series data.} \label{fig:td_vs_ctd}
\end{figure*}

When using the \gls{td} technique, an optimal dropout ratio value ($\alpha$) has to be identified to get a well-calibrated model.
This is a computationally expensive and time-consuming process, especially when working with over-parametrized models, commonly used in the \gls{eo} field and time series modeling.
In Gal et al. \cite{gal2017concrete} a continuous relaxation of the dropout technique is proposed, allowing the dropout ratio to be a learnable parameter in combination with standard gradient descent. 
The soft dropout mask $\tilde{\vec{p}}$ is defined as:
\begin{equation}
    \tilde{\vec{p}} = CS(\alpha, \vec{u}, \tau) =  \sigma \left( \left((\log{ \frac{\alpha}{(1 - \alpha)}} + \log{ \frac{\vec{u}}{(1- \vec{u})}} \right)  \frac{1}{\tau}  \right),
\end{equation}
with $\sigma$ the sigmoid function, $\tau \in \mathbb{R}^1$ a temperature value controlling the smoothness of the approximation, $\alpha \in \mathbb{R}^1$ a learnable parameter in the model, and $\vec{u}$ a uniform random variable, $u_i \sim \text{Uni}(0,1), \forall i = 1,...,T$. The latter acts as the auxiliary variable in the sampling reparametrisation \cite{maddison2017concrete}.
We use this sampling strategy in the dropout across time, named \acrfull{ctd}.
Furthermore, to get the uncertainty, we use this technique at inference time, referred to as \textbf{\acrfull{mcctd}}. This corresponds to obtaining $L$ samples of the Uniform distribution, $u_i^{(l)} \sim \text{Uni}(0,1), \forall l = 1, \ldots, L$, and forward over the model with data as $\hat{\vec{x}}^{(l)} = \vec{x} \odot (1-CS( \alpha, \vec{u}^{(l)}, \tau) )$. This technique is illustrated in Figure~\ref{fig:training}.
Additionally, in Figure~\ref{fig:td_vs_ctd} we schematically illustrate \gls{td} and \gls{ctd}. 

\section{Experiments} \label{sec:experiments}

\subsection{Datasets} \label{sec:experiments:data}

We use three publicly available \gls{eo} datasets with temporal sensor data. The characteristics of these datasets are presented in Table~\ref{tab:data}.
\paragraph{\gls{yield}}
We use a dataset for crop yield estimation \cite{perich2023pixel}, focused on cereals. 
This is a regression task in which the crop yield is estimated in tons per hectare (t/ha) from sensor data.
The data was collected in Switzerland between 2017 and 2021.
The temporal features are multi-spectral optical sensor (from Sentinel-2, with 10 bands) and weather data (with 5 bands).
The time series are from seeding to harvesting, with an approximate 5 days of sampling rate. 
All features were spatially interpolated to a pixel resolution of 10 m. 

\paragraph{\gls{lfmc}}
We use a dataset for moisture content estimation \cite{rao2020sar}. 
This involves a regression task in which the vegetation water (moisture) per dry biomass (in percentage) in a given location is predicted.
The data was collected between 2015 and 2019 in the western US.
The temporal features are multi-spectral optical sensor (from Landsat-8, with 8 bands) and radar sensor (from Sentinel-1, with 3 bands). 
These features were re-sampled monthly during a four-month window before the moisture measurement, i.e. a signal of 4 time-steps is used.
Additional static sensors are the topographic information, 
soil properties, 
canopy height, 
and land-cover class. 
All features were interpolated to a pixel resolution of 250 m.

\paragraph{\gls{pm25}}
We use a dataset for PM2.5 estimation \cite{pm25}. 
This involves a regression task in which the concentration of PM2.5 (particles measure 2.5 microns or fewer in diameter) in the air (in $\mu g/m^3$) in a particular city is predicted.
The data was collected between 2010 and 2015 in five Chinese cities.
The temporal features are atmospheric conditions (with 3 bands), atmospheric dynamics (with 4 bands), and precipitation (with 2 bands).
These are captured at hourly resolution, where we consider a three-day window for prediction. 

\begin{table}[!t]
    \centering
\begin{tabularx}{.9\textwidth}{>{\centering\arraybackslash}X>{\centering\arraybackslash}X    >{\centering\arraybackslash}X>{\centering\arraybackslash}X>{\centering\arraybackslash}X>{\centering\arraybackslash}X}
    \hline
    \textbf{Dataset} & \textbf{Samples} & \textbf{Series length} & \textbf{Features} & \textbf{Avg. target} & \textbf{Std. target} \\ 
    \hline
    \acrshort{yield} & 54,098 & $16$-$55$ & 15 & $7.356$ & $2.001$\\
    \acrshort{lfmc} & 2,578  & $4$ (fixed) & $61$ & $103.987$ & $39.562$ \\
    \acrshort{pm25} & 167,309 & $120$ (fixed)  & 9 & $73.673$ & $68.546$\\
    \hline
\end{tabularx}
    \caption{Datasets description and statistics.}     \label{tab:data}
\end{table}

\subsection{Setup} \label{sec:experiments:setup}

We use an early fusion strategy that concatenates all features along the time steps. For the architecture, we adopt a \gls{rnn} model with 2 layers based on long-short term memory units to extract the temporal information. Then, one additional layer is used to map the last hidden state of the \gls{rnn} into a hidden dimension. Finally, two prediction heads are employed, reflecting the mean and variance, respectively. All layers consist of 128 units, including 20\% of dropout. In addition, we use batch normalization layers after the \gls{rnn} model. 
We train the models for 100 epochs with an early stopping criteria based on a patience of five. The optimization is carried out with the ADAM optimizer and a batch size of 128 over the Mean Squared Error (MSE) function. 
For \gls{mc} sampling, 20 samples are used as in \cite{gal2016dropout}. We first evaluate \gls{mctd} and \gls{mcctd} against each other and then against standard \gls{uq} methods for time series regression tasks, including \gls{mc}-Dropout \cite{gal2016dropout} and a \gls{vi} \cite{blundell2015weight}. For comparison, all models employ the same architecture. If not differently specified, a \gls{td} ratio of 0.3 is used for the \gls{mctd} model, thereby aiming for a balance between regularization and model capacity.

\subsection{Results} \label{sec:experiments:results}

\begin{figure}[!t]
    \centering
    \includegraphics[width=\linewidth]{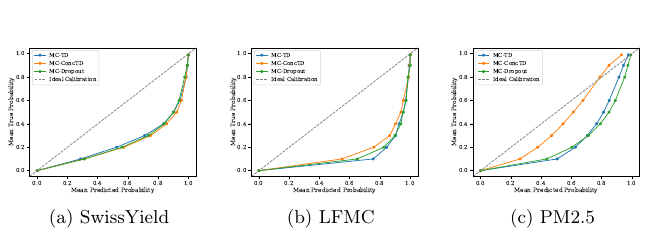}
    \caption{Model calibration across confidence levels for all datasets.} \label{fig:calibration_plots}
\end{figure}

\begin{table*}[!b]
    \centering
\begin{tabularx}{\textwidth}{cl|RRRRR}
\hline
\multicolumn{1}{l}{\textbf{Dataset}} & \textbf{Model} & \textbf{R2}($\uparrow$) & \textbf{RMSE}($\downarrow$) & \textbf{MAE}($\downarrow$) & \textbf{ECE}($\downarrow$) & {\textbf{PU}} ($\downarrow$)  \\ \hline
\multirow{2}{*}{SwissYield} & MC-TD     & $0.76$          & $0.99$           & $0.72$           & $\best{2.14}$ & $\best{1.00}$\\
                            & MC-ConcTD & $\best{0.79}$ & $\best{0.93}$  & $\best{0.66}$  & $2.23$  & $1.23$ \\ \hline
\multirow{2}{*}{LFMC}       & MC-TD     & $0.69$          & $21.92$          & $15.67$          & $3.77$ & $\best{1.01}$ \\
                            & MC-ConcTD & $\best{0.72}$ & $\best{20.80}$           & $\best{14.76}$          & $\best{3.41}$ & $8.01$ \\ \hline
\multirow{2}{*}{PM2.5}      & MC-TD     & $0.71$          & $36.36$          & $24.56$          & $2.92$ & $16.00$ \\
                            & MC-ConcTD & $\best{0.77}$ & $\best{31.77}$ & $\best{21.61}$ & $\best{2.15}$ & $\best{5.80}$ \\ 
\hline
\end{tabularx}

    \caption{Results of the two variants of dropout across time. The best results are highlighted in bold. The \gls{mctd} model uses a dropout ratio of $0.3$.}
    \label{tab:comp_TD}
\end{table*}
We compare the performance of \gls{mcctd} and \gls{mctd} models across all datasets in Table \ref{tab:comp_TD}. Both approaches demonstrate strong performance across all three datasets. However, \gls{mcctd} consistently outperforms  \gls{mctd} on all regression metrics for every dataset. For instance, a maximum improvement of 6 percentage points (p.p) is shown in $R^2$ on the \gls{pm25} dataset. \\
\begin{figure}[!b]
    \centering
    \includegraphics[width=\linewidth]{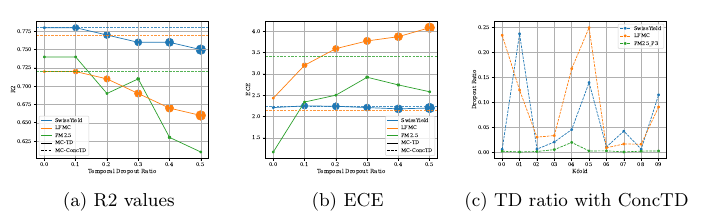}
    \caption{Model performance in $R^2$ and ECE under different TD settings, where marker size indicates the normalized PU.} \label{fig:TD_ratio_and_performance}
\end{figure}
In Figure \ref{fig:TD_ratio_and_performance} (a), we evaluate the performance of various dropout ratios in the \gls{mctd} model. The results indicate that a lower dropout ratio consistently improves performance across all datasets, with a clear decline in model performance as the ratio increases. Additionally, we observe an increase in PU with increasing \gls{td} ratio and decreasing performance. 
In Figure \ref{fig:TD_ratio_and_performance} (b) the calibration error (ECE) for every \gls{td} ratio is illustrated, together with the constant ECE for the \gls{mcctd} model. Marker sizes indicate normalized PU. Similarly, we observe an increase in calibration error with increased uncertainty, except for the \gls{yield} dataset,  and decreased performance. Ultimately, in Figure \ref{fig:TD_ratio_and_performance} (c), the learned dropout ratios of the \gls{mcctd} model across all folds are illustrated. Notably, the dropout ratios remain below 0.25 with significant instability across folds and datasets. For instance, approaching zero for the \gls{pm25} dataset. The results underline the difficulty of manually tuning the dropout ratio. \\
In Figure \ref{fig:calibration_plots}, we illustrate the model calibration. We define calibration as the models’ capabilities to accurately reflect the true probabilities of observed outcomes. For instance, for an 80\% confidence level, an event should occur approximately 80\% of the time. Ideally, a perfectly calibrated model aligns with the diagonal line.
In the \gls{yield} and \gls{lfmc} datasets, the models are overconfident, though the \gls{mcctd} model demonstrates relatively better calibration overall. Conversely, for the \gls{pm25} dataset, the models exhibit better calibration, especially for the \gls{mcctd} model. 
\begin{figure}[!t]
\centering
\includegraphics[width=0.7\linewidth]{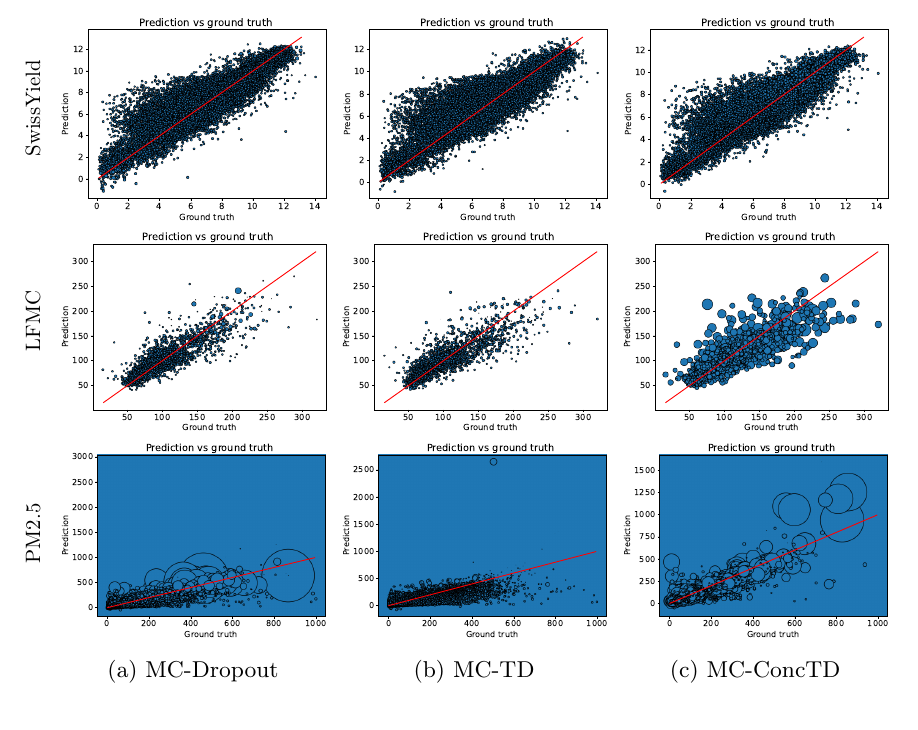}
\caption{Prediction over target plots for three regression datasets. The marker size indicates the uncertainty in each prediction.} 
\label{fig:scatterplot}
\end{figure}
In Figure \ref{fig:scatterplot}, prediction-versus-target plots are shown for \gls{mc}-Dropout, the \gls{mctd}, and \gls{mcctd} models on the three datasets. To analyze uncertainty estimates, the predictions are scaled based on the PU. Overall, the plots show a good alignment between predicted and target values. However, the distinct characteristics of the datasets become evident, with the \gls{pm25} dataset exhibiting very high uncertainties and the \gls{yield} dataset showing consistently low uncertainties. Furthermore, we observe that higher uncertainties are associated with lower alignment between predicted and target values, particularly for the \gls{mcctd} model.

\begin{table*}[!t]
    \centering
\begin{tabularx}{\textwidth}{cl|RRRRR}
\hline
\multicolumn{1}{l}{\textbf{Dataset}} & \textbf{Model} & \textbf{R2} ($\uparrow$)  & \textbf{RMSE} ($\downarrow$) & \textbf{MAE} ($\downarrow$)  & \textbf{ECE} ($\downarrow$) & {\textbf{PU}} ($\downarrow$)  \\ \hline
\multirow{4}{*}{SwissYield} & MC-Dropout                  & $0.78$          & $0.94$           & $0.67$           & $2.21$    & $1.02$      \\
                            & MC-TD                 & $0.76$          & $0.99$           & $0.72$           & $2.14$    & $\best{1.00}$      \\
                            & MC-ConcTD             & $\best{0.79}$ & $\best{0.93}$  & $\best{0.66}$  & $2.23$  & $1.23$         \\
                            & VI & $0.51$          & $2.02$           & $1.57$           & $\best{0.73}$ & $23.57$ \\ \hline
\multirow{4}{*}{LFMC}                & MC-Dropout           & $\best{0.72}$ & $\best{20.69}$ & $\best{14.64}$ & $\best{2.43}$ & $1.11$ \\
                            & MC-TD                 & $0.69$          & $21.92$          & $15.67$          & $3.77$  & $\best{1.01}$         \\
                            & MC-ConcTD             & $\best{0.72}$ & $20.80$           & $14.76$          & $3.41$    & $8.01$      \\
                            & VI & $0.53$          & $27.00$             & $20.00$             & $4.48$       & $4.01$   \\ \hline
\multirow{4}{*}{PM2.5}      & MC-Dropout                  & $0.74$          & $33.89$          & $22.45$          & $\best{1.17}$ & $3.99$  \\
                            & MC-TD                 & $0.71$          & $36.36$          & $24.56$          & $2.92$  & $16.00$         \\
                            & MC-ConcTD             & $\best{0.77}$ & $\best{31.77}$ & $\best{21.61}$ & $2.15$ & $5.80$          \\
                            & VI & $0.30$           & $57.00$             & $39.00$             & $3.57$      & $\best{2.49}$    \\ \hline
\end{tabularx}

    \caption{Performance comparison for various \gls{uq} models.} \label{tab:sota_uq}
\end{table*}
Finally, we compare the introduced models against two established \gls{uq} methods, namely \gls{mc}-Dropout and \gls{vi}.
The results are summarized in Table \ref{tab:sota_uq}.
Notably, the \gls{vi} performs poorly on regression metrics. 
The \gls{mcctd} model achieves the best R2 scores on the \gls{yield} and \gls{pm25} datasets, while being equal to the \gls{mc}-Dropout on the \gls{lfmc} dataset.
Considering all evaluation metrics, we find that the \gls{mcctd} has the best overall results.

\section{Discussion} \label{sec:discussion}

We empirically prove that \gls{td} at the inference level improves model performance over common \gls{uq} methods on various \gls{eo} regression tasks (Table~\ref{tab:comp_TD} and Table~\ref{tab:sota_uq}), thereby confirming related studies \cite{helber2024operational}.  
However, calibrating the dropout ratio in any \gls{dl} model is challenging as it requires exploring a continuous hyperparameter space. The same applies to the \gls{mctd} method. Consequently, \gls{mctd} can only find the optimal dropout ratio at high computational search costs.
Interestingly, higher dropout ratios or missing time-steps consistently result in reduced performance with increased uncertainty and calibration error, with an optimum value of around 0.1. The optimal value must balance predictive performance, uncertainty, and calibration error. This underlines the need for effective management of missing instances. Particularly, when time windows are small, determining the optimal ratio remains challenging.
Nevertheless, \gls{mcctd} demonstrates adaptability by learning different values based on the validation scenario (fold). This adaptability arises from the model’s ability to dynamically adjust the dropout value using the \gls{cd} \cite{gal2017concrete}, leading to improved performance and calibration. Nevertheless, we observe high variability across datasets. Moreover, we notice that there may be multiple optimal dropout ratios, as illustrated in Fig.~\ref{fig:TD_ratio_and_performance} (c). For instance, for the \gls{yield} dataset, the learned ratio of the second fold is approximately $0.25$, whereas in the sixth fold, it is approximately $0.01$. We attribute this variability to the distinct characteristics of each dataset, including varying time series length, number of features, and dataset size (Table~\ref{tab:data}). For instance, the \gls{lfmc} dataset is characterized by only four time steps, which could potentially explain the poor performance of \gls{td}-based models on this dataset. In contrast, the \gls{pm25} dataset has 120 time steps, but has only short temporal dependencies compared to the \gls{yield} dataset. Therefore, dropping the previous time-steps can significantly reduce performance. As a result, \gls{td} is less effective in this context, potentially explaining the low \gls{td} ratios in Fig.~\ref{fig:TD_ratio_and_performance} (c).
Overall, \gls{ctd} enhances the robustness of UQ across different EO datasets, by continuously adapting to unique data characteristics while reducing the computational burden of optimal ratio search, making \gls{uq} more accessible for \gls{eo} applications with time series data. \\

\textbf{Limitations}
The proposed methods and experimental setup have limitations that must be considered. 
We validate the models using \gls{eo} data without real missing values in the time series, which could potentially differ from scenarios involving actual missing data. This limitation may affect how the models perform in real-world situations where missing values are present.
Additionally, we use only an \gls{rnn} encoder to learn the temporal patterns and only few \gls{uq} methods for comparison. However, the primary goal of this study is to demonstrate the effectiveness of \gls{td} and \gls{ctd} in enhancing predictive performance and \gls{uq}, rather than identifying the optimal architecture for each dataset and use case. 
Nevertheless, additional models and architectures must be evaluated in the future, including Transformers, and other \gls{uq} methods. 
Moreover, while \gls{td} has shown significant improvements in related literature \cite{helber2024operational}, we observe only small improvements in this study on individual datasets. This may be attributed to the task and the unique data characteristics, including short time series length (\gls{lfmc}), short temporal dependencies (\gls{pm25}), or noise in remote sensing data. More datasets must be considered to better understand the robustness of the proposed methods. 
Finally, while we demonstrate improvements in our models regarding the literature, we emphasize that poor calibration remains an ongoing challenge that requires careful evaluation, an issue commonly encountered in \gls{dl} \cite{guo2017calibration,valdenegro2021find}. Specifically, \gls{eo} data is often impacted by noisy und uncertain measurements and spatio-temporal distribution shifts, which can lead to poor calibration when the model is applied to unknown environments. Further exploring model calibration, including post-hoc calibration, will be required before deploying models into practice.

\section{Conclusion} \label{sec:conclusion}

This work underscores the importance of input uncertainty in time-series data for \gls{eo} regression applications. 
To address this, we introduced two novel uncertainty estimation methods, namely \gls{mctd} and \gls{mcctd}. Both methods account for input uncertainty in time series data by applying \acrfull{td} at the inference level using Monte Carlo sampling. While \gls{mctd} requires manual and expensive tuning of the dropout ratio, \gls{mcctd} learns this automatically as a free parameter. We empirically demonstrate their effectiveness in enhancing predictive performance in various \gls{eo} datasets. While these models improve the accessibility of uncertainty estimation in \gls{eo} applications, they also present challenges that require further investigation.
Future research will focus on validating these approaches across diverse \gls{eo} applications and by considering model calibration.

\begin{credits}
\subsubsection{\ackname} We acknowledge the support of the University of Kaiserslautern-Landau (RPTU) and the German Research Center for Artificial Intelligence (DFKI).
\subsubsection{\discintname}
The authors have no competing interests to declare that are relevant to the content of this article.     
\end{credits}

\bibliographystyle{splncs04}
\bibliography{main}

\end{document}